\documentclass[referee, sn-mathphys-num]{sn-jnl}

\usepackage{graphicx}%
\usepackage{multirow}%
\usepackage{amsmath,amssymb,amsfonts}%
\usepackage{amsthm}%
\usepackage{mathrsfs}%
\usepackage[title]{appendix}%
\usepackage{xcolor}%
\usepackage{textcomp}%
\usepackage{manyfoot}%
\usepackage{booktabs}%
\usepackage{algorithm}%
\usepackage{algorithmicx}%
\usepackage{algpseudocode}%
\usepackage{listings}%
\usepackage{subfig}
\usepackage{graphbox}
\usepackage{enumitem}
\usepackage{wrapfig}
\usepackage{lineno}

\begin{document}
\title[Article Title]{Network-Based Transfer Learning Helps Improve Short-Term Crime Prediction Accuracy}



\author[2]{\fnm{Jiahui} \sur{Wu}}

\author[1]{\fnm{Vanessa} \sur{Frias-Martinez}}

\affil[1]{
\orgname{University of Maryland}}

\affil[2]{ \orgname{Huawei}}



\abstract{
Deep learning architectures enhanced with human mobility data have been shown to improve the accuracy of short-term crime prediction models trained with historical crime data.
However, human mobility data may be scarce in some regions, negatively impacting the correct training of these models. 
To address this issue, we 
propose a novel transfer learning framework for short-term crime prediction models, whereby weights from the deep learning crime prediction models trained in \textit{source} regions with plenty of mobility data are transferred to \textit{target} regions to fine-tune their local crime prediction models and improve crime prediction accuracy.
Our results show that the proposed transfer learning framework improves the F1 scores for target cities with mobility data scarcity, especially when the number of months of available mobility data is small. We also show that the F1 score improvements are pervasive across different types of crimes and diverse cities in the US. 
}




\maketitle


\section{Introduction}

Crimes negatively impact the wellbeing of individuals and society as a whole. 
Researchers from various fields such as criminology, geographic information science, urban planning and data science, have conducted studies about the patterns of urban crimes. These studies help us better understand when and why certain crimes might happen and, more importantly, provide insights into the design of interventions to reduce the volumes of crimes.
One critical research direction of such efforts is place-based crime prediction that focuses on predicting the number of crime incidents or crime occurrence for a given location.
Through the lens of place-based crime prediction, we can study the complex relationship between future crimes and historical crimes, the built environment and social interactions in different places. 
Place-based crime predictions are typically carried out using either long-term or short-term approaches. 
Long-term crime prediction analysis, such as monthly or annual crime prediction, allows us to understand how the environmental factors of places shape future crimes; and in turn, help us inform better urban planning that improves the urban environment potentially decreasing crime occurrence. On the other hand, short-term crime prediction analysis focuses on next-day crime prediction {\it i.e.,} the identification of places where there will be crimes the next day. Short-term crime prediction is generally used to better allocate policing resources to response to crimes more swiftly. 
In this paper, we focus on short-term crime prediction analysis.

Various models have been developed to tackle this problem. From kernel density estimation 
that uses the estimated density of historical crimes as a measure of risk for future crime areas  
\cite{Chainey2008-ys}, to  
epidemiological models 
whereby the spatio-temporal patterns of crimes in one location increase the probability of other incidents occurring at nearby locations \cite{Johnson2014-fc, Mohler2015-gm}.
Nevertheless, more recent deep learning approaches have shown superior performance in short-term crime prediction by modeling the spatio-temporal patterns of crime data in the built environment as non-linear patterns \cite{Duan2017-fg, Huang2019-sx, Wu2020-ml}.
Based on the \textit{crime opportunity theory} that suggests that human mobility is a key factor in crime generation \textit{i.e.}, the higher the presence of people or property, the more crimes could happen \cite{hannon2002criminal}, recent work has also shown that deep learning architectures enhanced with mobility data characterizing local mobility patterns can improve further the accuracy of the short-term crime prediction models trained with historical crime data \cite{wu2022enhancing}.

The rise of information and communication technologies (ICT) such as mobile phones and wearable devices, and location-based services \textit{e.g.}, geotagged social media or ride sharing, has generated large amounts of human mobility data in cities with well developed infrastructures.
However, human mobility data may be scarce in some regions, including certain rural areas or regions in less-resourced countries, where the infrastructure and mobility services might be limited or lacking. 
As a result, the mobility data collected in these regions might be insufficient to properly train deep-learning, short-term crime prediction models. 

In this paper, we propose to address the lack of mobility data with a novel transfer learning framework for deep learning crime prediction models. In this context, transfer learning aims to extract knowledge from one or more \textit{source} regions where mobility data is available and apply that knowledge to a \textit{target} region with limited mobility data with the main objective of improving the accuracy of deep learning, short-term crime prediction models \cite{Pan2010-kk}.
Building on prior work that has shown promising results of transfer learning methods for traffic and flow prediction \cite{Wang2019-ie, Yao2019-ie}, the proposed transfer learning framework is framed as a cross-region transfer learning whereby short-term crime prediction knowledge from source regions is leveraged to improve crime prediction models for the target regions \cite{Wang2019-ie, Yao2019-ie}.
Specifically, we use a network-based transfer learning approach whereby weights from the deep learning crime prediction models trained in \textit{source} regions with plenty of mobility data are transferred to \textit{target} regions to fine-tune their local crime prediction models and improve crime prediction accuracy \cite{Tan2018-kq}.


To evaluate whether the proposed transfer learning framework for deep learning, short-term crime prediction models helps regions with 
limited mobility data improve crime prediction accuracy by leveraging knowledge from regions with abundant mobility data, 
we use publicly available crime data as well as publicly available fine-grained human mobility data from a large-scale mobile phone dataset in the US \cite{kang2020multiscale}. 
For evaluation purposes, we frame regions as cities, and focus our analysis on crime prediction transfer knowledge across four American cities (Austin, Baltimore, Chicago and Minneapolis) and for multiple types of crimes. 
The main contributions of this paper are:

\begin{itemize}
    \item A novel transfer learning framework to improve the accuracy of deep learning, short-term crime prediction models when mobility data is scarce. Our results show that the proposed approach improves the prediction accuracy, especially when the number of months for which mobility data is available is low. 
    \item A thorough analysis of the transfer learning framework across four cities, eight types of crimes and diverse levels of data scarcity, showing that accuracy improvements are pervasive across diverse contexts and highlighting differences. 
\end{itemize}

\section{Related Work}

\subsection{Crime Prediction Models and Mobility Data}
Historical crime data and socioeconomic data are often used in crime prediction models \cite{Catlett2019-ra}.
For example, historical crime hotspots have been used to assess the risk of future crimes 
\cite{Catlett2019-ra}; and the relationship of dependency between different types of crimes has
also been used to predict future crimes \cite{Mohler2015-gm}
Neural networks have also been applied to modeling spatiotemporal patterns in historical crimes for future crime prediction \cite{Wang2017-ia, Yao2019-ie, Wu2020-ml}. In the neural networks, the spatial patterns of crimes are modeled by convolution layers, while the temporal patterns can either be modeled as multiple feature maps in the convolution layers \cite{Wang2017-ia} or modeled by the recurrent neural network layers such as LSTM \cite{Yao2019-ie}.

In addition to historical crimes, census data \cite{Kadar2018-ku}, points of interest (POI) \cite{Zhao2017-da} or mobility data \cite{wu2022enhancing} have been shown to enhance crime prediction models. 
Mobility data for crime prediction is often modeled as an origin-destination matrix (OD) that characterizes human mobility (flows) between census tracts. Human OD mobility data has been used to characterize human behaviors in the built environment 
~\cite{vieira2010querying,hernandez2017estimating,frias2013cell,rubio2010human,wuspatial}, 
for public safety~\cite{wu2022enhancing,wu2023auditing}, during epidemics and 
disasters~\cite{wesolowski2012quantifying,bengtsson2015using,hong2017understanding,isaacman2018modeling,ghurye2016framework,hong2020modeling, Erfani_Frias-Martinez_2023,abrar2023analysis}, as well as to support decision making for socio-economic development
~\cite{frias2010socio,fu2018identifying,frias2012mobilizing, hong2016topic,frias2012computing,hong2019characterization}.
In this paper, we will focus on deep learning crime prediction models that exploit the predictive power past crime data and OD mobility matrices \cite{wu2022enhancing}.

\subsection{Transfer Learning}
Transfer learning is important when training data is insufficient. Generally speaking, transfer learning aims to extract the knowledge from one or more source settings (tasks) and to apply that knowledge to a target setting (task) \cite{Pan2010-kk}. In the context of urban computing, cross-city transfer learning aims to transfer knowledge from source cities with abundant data resources to target cities where services and infrastructures are not ready or just in place, and where data resources are insufficient.
Cross-city transfer learning often times is described as domain adaptation, as in Pan and Yang's framework, where the tasks are the same for both source and target cities \cite{Pan2010-kk}. 
Cross-city transfer learning has been applied to multiple areas in urban computing including POI recommendation in new cities\cite{Xu2018-uo, Ding2019-cz}, mobility generation \cite{Fan2016-gw, He2020-sz}, bike services distribution \cite{Liu2018-go}, crowd flow prediction \cite{Wang2019-ie, Yao2019-ie, Mallick2020-hm}.

In this context, there exist four categories of transfer learning for deep learning models \cite{Tan2018-kq}: 
a) instances-based, which utilizes instances in the source domain by assigning appropriate weights; 
b) mapping-based, which maps instances from two domains into a new data space with better similarity;
c) network-based, which reuses parts of a pre-trained network in the source domain; and
d) adversarial-based, which uses adversarial technology to find transferable features suitable for the two domains. 
In this paper, we focus on network-based approaches to transfer knowledge from source, rich-data cities to target, poor-data cities; and explore whether that knowledge transfer improves the accuracy of short-term crime prediction models.

\section{Data}
\label{s3-sec:Dataset}
To evaluate the feasibility of transferring crime prediction knowledge from regions with large mobility data availability to regions with limited mobility data, we will define regions as cities, and we will retrieve crime incidents and human mobility data for each of the cities in this study. In this paper, we focus on Baltimore (Bal), Minneapolis (Min), Austin (Aus) and Chicago (Chi). 
These four cities were chosen based on the diversity of their demographics, as shown in Table \ref{s3-tab:race}, with Baltimore having majority Black and African-American population, Minneapolis majority White, Austin has a high White and Latino and Hispanic population and Chicago with a balanced mix of White, Black and African-American and Hispanic and Latino communities. Replicating the short-term crime prediction and fairness analysis across these four cities will provide a robust analysis across geographies.

\begin{table}[]
\small
\centering
\begin{tabular}{cp{1.5in}p{1.5in}p{0.8in}p{0.8in}}
\toprule
    & \%   Not Hispanic or Latino, White Alone& \%   Black or African-American & \%   Hispanic or Latino & \% Asian \\
\midrule
Bal & 27.54\%                           & 62.46\%                & 5.12\%                   & 2.59\%   \\
Min & 59.80\%                           & 19.36\%                & 9.58\%                   & 6.13\%   \\
Aus & 49.08\%                           & 7.60\%                 & 33.64\%                  & 7.34\%   \\
Chi & 33.61\%                           & 29.48\%                & 28.89\%                  & 6.40\%  \\
\bottomrule
\end{tabular}
\caption{The percentage of population across race and ethnicity for the four cities according to the American Community Survey (2019 ACS 5-year estimates)\cite{us_acs}. The cities are: Baltimore (Bal), Minneapolis (Min), Austin (Aus) and Chicago (Chi).}
\label{s3-tab:race}
\end{table}

\subsection{Crime incident data} 
The crime incident datasets for the four cities are obtained from their open data portals, covering crimes from January to December, 2020\footnote{ 
Bal: https://data.baltimorecity.gov/; 
Min: https://opendata.minneapolismn.gov/; \newline
Aus: https://data.austintexas.gov/; 
Chi: https://data.cityofchicago.org/;
}. 
Each crime incident is associated with the crime category it belongs to and with the time and location where it took place. Crime locations are generally geo-coded to the closest street or block in the city, however, to account for the potential spatial precision inaccuracy, 
We use a 50-meter buffer to associate crime incidents to urban census tracts (a similar approach has been implemented 
in prior work {\it e.g.,} \citet{Kadar2018-ku,De_Nadai2020-gt}).
Although crime incidents could be associated to smaller spatial units, the choice for spatial units is determined by the availability of human mobility data at the census tract level only. 
We group the crime incidents into two types: property and violent crimes, and we will evaluate the transfer learning framework for each type separately. Property crimes include arson, burglary, larceny-theft, and motor vehicle theft; while violent crimes include aggravated assault, forcible rape, murder, and robbery.
Tables \ref{s3-tab:crm_density} and \ref{s3-tab:crm_density1} show the monthly crime density for each city throughout 2020, where monthly crime density is computed as the percentage of census tracts with crime incidents during that month. 
The table shows that the four cities selected generally suffer from higher volumes of property crimes than violent crimes; and that they represent a diverse group with some cities suffering from higher volumes of violent and property crimes than others.

\begin{table}[]
\small
\centering
\begin{tabular}{@{}llllllll@{}}
\toprule
         &     & Jan    & Feb    & Mar    & Apr    & May    & Jun       \\ 
\midrule
         & Bal & 28.0\% & 27.2\% & 24.6\% & 22.4\% & 23.6\% & 25.0\% \\
Property & Min & 35.0\% & 33.4\% & 34.1\% & 35.3\% & 37.6\% & 34.7\% \\
Crime    & Aus & 32.9\% & 31.9\% & 30.6\% & 30.5\% & 31.2\% & 31.5\% \\
         & Chi & 23.5\% & 22.6\% & 19.7\% & 16.6\% & 19.6\% & 20.4\%  \\
\midrule
         & Bal & 21.6\% & 21.1\% & 21.8\% & 17.0\% & 21.6\% & 23.4\%  \\
Violent  & Min & 9.4\%  & 9.3\%  & 10.7\% & 8.5\%  & 10.3\% & 13.0\%   \\
Crime    & Aus & 4.0\%  & 3.7\%  & 4.5\%  & 4.2\%  & 5.0\%  & 5.4\%    \\
         & Chi & 11.5\% & 11.0\% & 9.9\%  & 8.3\%  & 10.2\% & 11.6\%  \\ 
\bottomrule
\end{tabular}
\caption{Crime occurrence monthly density for the four cities in 2020 (Jan-June): Baltimore (Bal), Minneapolis (Min), Austin (Aus) and Chicago (Chi).}
\label{s3-tab:crm_density}
\end{table}

\begin{table}[]
\small
\centering
\begin{tabular}{@{}llllllll@{}}
\toprule
         &      & Jul    & Aug    & Sep    & Oct    & Nov    & Dec    \\ 
\midrule
         & Bal  & 24.0\% & 22.7\% & 24.7\% & 25.6\% & 24.2\% & 21.3\% \\
Property & Min  & 41.6\% & 43.3\% & 40.7\% & 41.3\% & 37.0\% & 33.2\% \\
Crime    & Aus  & 31.8\% & 34.3\% & 35.0\% & 33.3\% & 36.1\% & 34.4\% \\
         & Chi  & 22.5\% & 23.5\% & 22.2\% & 21.0\% & 19.7\% & 18.2\% \\
\midrule
         & Bal  & 23.2\% & 23.4\% & 22.4\% & 22.5\% & 21.1\% & 18.6\% \\
Violent  & Min  & 16.4\% & 14.6\% & 13.7\% & 12.9\% & 10.4\% & 8.3\%  \\
Crime    & Aus   & 5.7\%  & 5.3\%  & 5.2\%  & 4.7\%  & 5.3\%  & 5.2\%  \\
         & Chi  & 12.9\% & 12.8\% & 12.4\% & 11.1\% & 10.8\% & 9.3\%  \\ 
\bottomrule
\end{tabular}
\caption{Crime occurrence monthly density for the four cities in 2020 (July-Dec): Baltimore (Bal), Minneapolis (Min), Austin (Aus) and Chicago (Chi).}
\label{s3-tab:crm_density1}
\end{table}

\subsection{Human mobility data}
The pervasive presence of ubiquitous technologies such as smart phones, has allowed for the collection of large-scale human mobility data. Location intelligence companies like SafeGraph, collect pseudonymized mobile GPS location data using SDKs installed on individuals' mobile phones via mobile apps. SafeGraph offers multiple datasets. For this study, we have used daily origin-to-destination flows at the census tract (CT) level from January to December, 2020. This dataset is publicly available (see \cite{kang2020multiscale}). To extract this dataset, SafeGraph assigns to each device a home location at the census block group level based on its night-time activity. Then, it tracks for each device all the trips from its home location to points-of-interest (POIs) in SafeGraphs' large POI database. Origin-destination (OD) flows are finally computed by transforming all the home-to-POIs trips to CT(O)-CT(D) trips and by computing the number of devices associated to each OD across all census tracts in a city. OD flow volumes are computed at a daily granularity. Since the devices in SafeGraph's database account for about 10\% of the entire population in the U.S., the OD flow volumes are re-scaled by the census population.

\begin{table}[]
\small
\centering
\begin{tabular}{p{1.8in}p{0.8in}p{0.8in}p{0.8in}p{0.8in}}
\toprule
 & Bal & Min & Aus & Chi \\
\midrule
Number of census tracts & 200 & 116 & 204 & 809 \\
\midrule
Volume of in-city OD flow & 4040.1 (1733.9) & 4004.3 (1653.7) & 8167.2 (3866.3) & 5307.3 (2821.6) \\
\midrule
Volume of out-of-city OD flow  & 1413.6 (1149.9) & 2055.8 (1749.5) & 2102.6 (1651.3) & 1198.9 (1646.3) \\
\midrule
The number of unique census tracts connected by in-city OD flow  & 38.7 (14.6)     & 30.5 (10.7)     & 66.8 (20.2)     & 61.0 (28.5)     \\
\midrule
The number of unique counties connected by out-of-city OD flow & 14.5 (11.9)       & 23.6 (20.8)       & 29.6 (17.2)       & 15.1 (20.5)      \\
\midrule
The number of unique states connected by out-of-city OD flow & 5.9 (3.3)       & 7.0 (4.2)       & 7.7 (4.0)       & 6.2 (4.0)      \\
\bottomrule
\end{tabular}
			
\caption{Human mobility flow statistics for the four cities under study: Baltimore (Bal), Minneapolis (Min), Austin (Aus) and Chicago (Chi). The numbers in each cell represent the mean (standard deviation) of the daily average across all census tracts in a given city in 2020. OD flows outside the city are flows that either start or end in a census tract that is not part of the city of interest.} 
\label{s3-tab:mobi_stats}
\end{table}

Table \ref{s3-tab:mobi_stats} shows general OD flow volume statistics for the four cities under study for the year 2020.
For each measure, the table shows the mean and standard deviation of its daily average values across all census tracts in each city.
In-city OD flows refer to flows whose origin and destination census tracts (CT(O) and CT(D)) are within the city; while out-of-city OD flows are flows in which either the origin or the destination census tract is outside the city under study.
To characterize mobility diversity, the table also shows the number of unique census tracts connected by in-city OD flows and the number of unique counties and states connected by out-of-city OD flows. 
It can be observed that most of the OD flows identified take place within the cities under study, with smaller volumes being associated to trips to counties outside the city, and even a smaller number to trips to other states. Consequently, there is a higher diversity in the number of distinct areas visited inside than outside the city. 
A more detailed description of the features extracted from this dataset is covered in the next section.


\section{Short-term Crime Prediction with Mobility Data}
Our main objective is to evaluate whether transfer learning can help regions with 
limited mobility data improve the accuracy of short-term crime prediction models by leveraging knowledge from regions with abundant mobility data.
In this section, we describe the short-term crime prediction in detail, and in the next section we will present the transfer learning framework.

\subsection{Problem setting}
\label{s3-sec:problem-setting}
In this study, we focus on placed-based short-term crime prediction for a given city. For that purpose, 
a city is divided into $N$ spatial units $\mathbf{S}=\{s_1, s_2, ..., s_N\}$ which for this study are defined as census tracts. Census tracts are chosen as spatial units because the human mobility flow dataset is only available at the census tract level. 
The short-term crime prediction is framed as determining whether there will be at least one crime the next day at a given census tract using prior crime and mobility data for that tract. 
Crime occurrences at a census tract $s_i$ on day $t$ are denoted as $h_{i,t}$ and $h_{i,t}=1$ is referred to as a crime hotspot. 

For each census tract $s_i$, two sets of daily predictive features are computed: 
1) historical crimes ($C$), defined as the daily number of past crime incidents; the input sequence for crime prediction at day $t$ is represented as
$\mathbf{C}_{i,t}=\{c_{i,t-T}, c_{i, t-T+1}, ..., c_{i, t-1}\}$ with  
$T$ being the length of the \textit{look-back} period {\it i.e.,} the time range used to characterize {\it history} and $c_{i, t-d}$ being the number of crime incidents $d$ days before day $t$;
and 
2) mobility features ($M$), defined as a set of ten daily features extracted from SafeGraph's daily OD matrices and denoted as
$\mathbf{M}_{i,t} = \{\mathbf{M}_{i,t}^{j} | j \in \{1, 2, ..., 10\}\}$ and $\mathbf{M}_{i,t}^{j} = \{m_{i,t-T}^{j}, m_{i, t-T+1}^{j}, ..., m_{i, t-1}^{j}\}$, where $m_{i,t-d}^{j}$ is the value of the $j$-th mobility feature at $d$ days before day $t$.
The ten features identified characterize mobility volumes and mobility diversity. 
Mobility volume features characterize the daily total number of people going in (inflow) and out (outflow) of a census tract within or outside the city under study, which have been shown to be related with the volumes of crime incidents \cite{Kadar2018-ku, Bogomolov2015-pd, Wu2020-ty};
while mobility diversity features characterize the regional influence, \textit{i.e.}, the number of unique regions visited by in/outflows, including census tracts, counties and states. Past research has shown that crimes committed by visitors are associated to different patterns (behaviors) than those of residents \cite{Boivin2018-cf}; and that pass-through traffic information improves crime prediction accuracy \cite{Kadar2020-ij}. Therefore, mobility diversity features are extracted to reflect the connections between the census tract $s_i$ and other regions.
Table \ref{s3-tab:ftr} shows a summary of all the features used in the short-term crime prediction models. Besides crime and human mobility data, we also add {\it Day of week} to the feature set to capture the difference between crime data and human mobility behaviors during weekdays and weekends.

\begin{table}[]
\centering
\small
\begin{tabular}{@{}ll@{}}
\toprule
Types     & Features\\
\midrule
Crimes    & Daily number of crimes\\
\midrule
Mobility  & Volumes of in-city inflow                   \\
Volumes   & Volumes of in-city outflow                   \\
          & Volumes of out-of-city inflow           \\
          & Volumes of out-of-city outflow             \\
\midrule
Mobility  & Number of CT connected by in-city inflow             \\
Diversity & Number of CT connected by in-city outflow            \\
          & Number of counties connected by out-of-city inflow  \\
          & Number of counties connected by out-of-city outflow \\
          & Number of states connected by out-of-city inflow    \\
          & Number of states connected by out-of-city outflow     \\
\midrule
day of week & Day of week \\
\bottomrule
\end{tabular}
\caption{Complete list of predictive (input) features for short-term crime prediction models. For census tract $s_i$, inflow (outflow) means $s_i$ is the destination (origin) of the OD flow.}
\label{s3-tab:ftr}
\end{table}

\textbf{Problem Statement}. Given the temporal sequences of input features ($C+M$) within the \textit{look-back} period $T$ for all census tracts in a city,
predict whether a census tract will be a hotspot (or not) in the next day $h_{i,t}=1, i\in[1,N]$ or $h_{i,t}=0$, otherwise.


\subsection{Deep Learning Model}
\begin{figure}
    \centering
    \includegraphics[width=.45\linewidth]{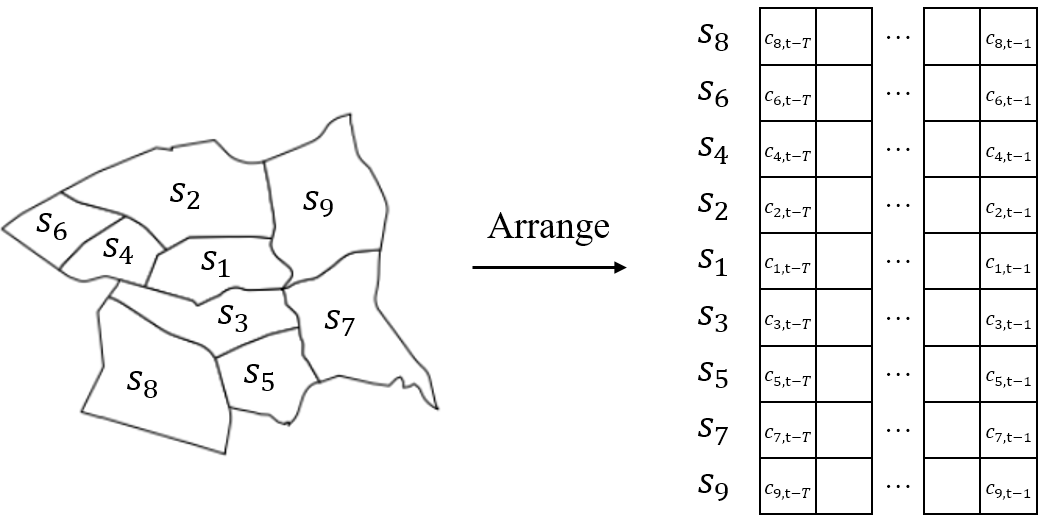}
    \caption{Arrange the nearest neighbors set for the target census tract $s_1$ and construct the 2D feature map for historical crimes. In the neighboring set of $s_1$, $s_2$ and $s_3$ is the closest to $s_1$; $s_4$ and $s_5$ are the next closest to $s_2$ and $s_3$ respectively; $s_6$ and $s_7$ are the next closest to $s_4$ and $s_5$; $s_8$ and $s_9$ are the next closest to $s_6$ and $s_7$. Similar process is applied to each of the ten mobility features.}
    \label{s3-fig:nbcnn_rearrange}
\end{figure}

Neighbor convolution models (NbConv) that account for spatio-temporal dependency have been used for crime prediction using historical data over a spatial grid \cite{Duan2017-fg}. NbConv models have also been used for crime prediction using both historical crime and mobility data, and prior work has shown a superior predictive performance of NbConv when compared with other deep learning approaches  
\cite{wu2022enhancing}. As a result, we select NbConv as our short-term crime prediction model. 
To adapt the NbConv model to the  setting in this study, where the spatial units are census tracts (non-regular division), we extract a fixed-length nearest neighbors set for each census tract for which the model outputs the next-day crime prediction. Specifically, we focus on the eight nearest census tracts for a given census tract. 
We arrange the target census tract in the middle and sort the nearest neighboring census tracts from closest to furthest to form a 2D feature map per input feature, as explained in Figure \ref{s3-fig:nbcnn_rearrange}. Such arrangement allows the kernel of the convolutional layer to model the spatio-temporal dependency through its local receptive field. These 2D feature maps are then input to the full convolution architecture.

\section{Transfer Learning Framework}
Our objective is to evaluate the feasibility of transferring crime prediction knowledge from source regions with large mobility data availability to target regions with limited mobility data so as to improve the short-term crime prediction accuracy of NbConv models. 
Given that both source and target models share the same network structure (NbConv), we frame transfer learning as network-based 
distilling knowledge from the data distribution in the source city in the form of learned parameters from the neural network NbConv model. 

\begin{figure}
    \centering
    \includegraphics[width=\linewidth]{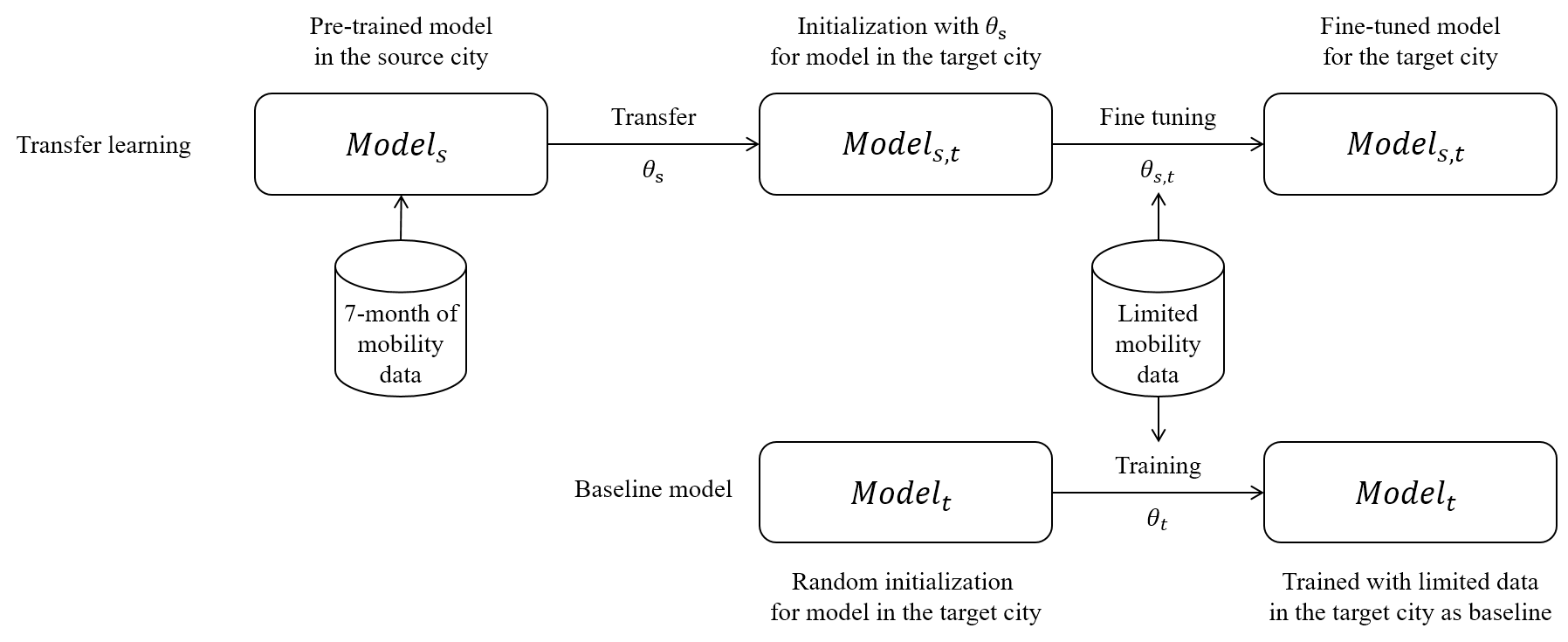}
    \caption{The framework of the transfer learning technique applied in this study. 
    $Model_s$ and $Model_{s,t}$ have the same network architecture. The parameters $\theta_s$ of the whole architecture of $Model_s$ is transferred to $Model_{s,t}$ as the initialization of $\theta_{s,t}$.
    }
    \label{s3-fig:transfer-learning}
\end{figure}

Figure \ref{s3-fig:transfer-learning} shows the main characteristics of the framework proposed. The proposed transfer learning follows these steps:
1) the source model is pre-trained with the data in the source city, denoted as $Model_s$;
2) the learned parameters (weights) of all layers in the whole architecture of $Model_s$, \textit{i.e.}, $\theta_s$, are used to initialize the parameters of model for the target city $Model_{s,t}$;
3) the parameters of $Model_{s,t}$ are fine-tuned with the limited data in the target city. Fine-tuning refers to the process of updating the transferred parameters $\theta_s$ with the local limited data in the target city and the post-fine-tuning parameters are denoted as $\theta_{s,t}$. $Model_{s,t}$ will be also referred as a fine-tuned model.

Given the 1 year of crime and mobility data, we 
chronologically split the dataset into training (6.5 months), validation (0.5 month), and testing (5 month) sets. 
The validation set is used to tune the learning rate and early stopping {\it i.e.,} deciding the maximum number of epochs for training. 
For each testing month, $Model_s$ is pre-trained with training data from the previous 7 months in the source city. 

To simulate different levels of data scarcity, we vary the number of months of "limited mobility data" in the target city from 1 to 7 months. The shorter the length of data available in the target city is, the more severe the data scarcity issue is for the target city. 
7 months of mobility data (which is as long as the training data in the source city) is included in the experiment to evaluate the effects of transfer learning when the target city also has abundant mobility data.  
The overall performance of a model in the framework is represented by its monthly F1 score, computed comparing the next-day crime prediction with the daily ground truth over all days for each testing month (August to December, 2020).

\subsection{Transfer Learning Evaluation Protocol}

In order to evaluate whether transfer learning can improve crime prediction accuracy in the target city, a baseline model for the target city $Model_t$ for each level of data scarcity (number of months of collected mobility data) is trained as a random-initialized model with limited mobility data, as shown in Figure \ref{s3-fig:transfer-learning} (bottom). In order words, a baseline model is trained in the same way as the models in the source cities, but with a shorter length of training data (from only 1 to all 7 months) to simulate data scarcity.
Next, we use the relative change in the average monthly F1 score between the baseline model and a fine-tuned model using knowledge transferred from a data-rich source city to evaluate the effect of transfer learning: $\frac{F1_{model_{s,t}}}{F1_{model_t}} - 1 * 100\%$. Positive (negative) relative change in F1 score suggests that the transferred knowledge from the source city improves (degrades) the prediction accuracy when compared with the baseline model, trained only with limited data in the target city.

To holistically evaluate the effects of transfer learning in crime prediction accuracy, we treat each of the four cities as data-scarce target cities with the remaining three being treated as the data-rich source cities. For example, when Baltimore is considered a target city, Minneapolis, Austin and Chicago are considered source cities from which knowledge can be extracted.
Our experimental evaluation considers both a unique source city for transfer learning \textit{e.g,} transfer learning from Austin to Minneapolis, as well as multi-city transfer learning whereby knowledge from all source cities is transferred. 
To make use of transferred knowledge from multiple source cities, we apply a majority voting to aggregate the next-day crime predictions from multiple fine-tuned models for a target city, \textit{i.e.}, a census tract in the target city is predicted as a crime hotspot in the next day if it is predicted as a hotspot by at least half of the fine-tuned models. For example, when Baltimore is the target city, there are three fine-tuned models with transferred knowledge from Minneapolis, Austin and Chicago. A census tract in Baltimore is considered as a hotspot in the next day if it is a predicted hotspot in at least two of the fine-tuned models.

\section{Results}

\begin{figure}
    \centering
    \includegraphics[width=\linewidth]{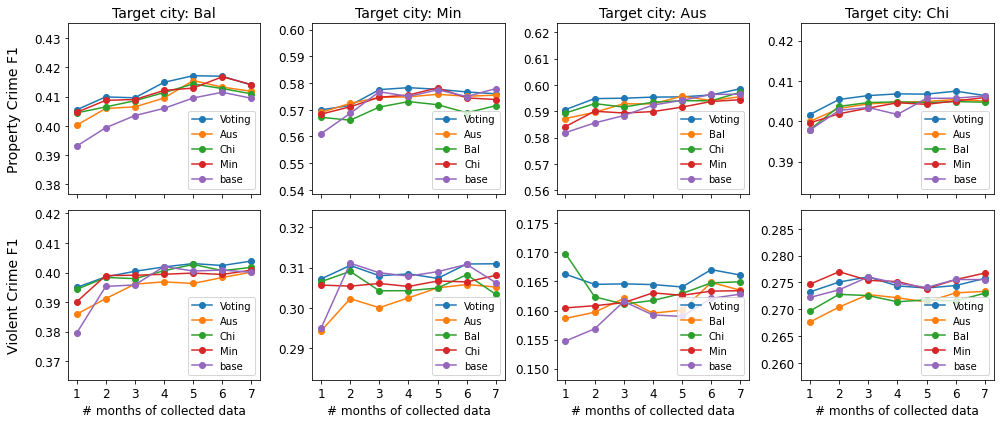}
    \caption{Average monthly F1 score for crime prediction by fine-tuned models with transferred knowledge from different source cities. }
    \label{s3-fig:transfer-f1}
\end{figure}

Figure \ref{s3-fig:transfer-f1} shows the average monthly F1 score of the fine-tuned crime prediction models with transferred knowledge from different source cities and of the baseline model without transfer learning in each target city. 
Each column shows the results for property and violent crime prediction in each target city. In each plot, the x-axis is the number of months of collected mobility data in the target city; \textit{voting} refers to the majority voting aggregating knowledge from multiple source cities, each of the three cities refer to the fine-tuned model with data from that source city; and \textit{base} refers to the baseline model with training data from the target city only.

The \textit{baseline} (purple) lines in all plots suggest that across all cities and types of crimes, the prediction F1 score tends to be smaller when the number of months of collected data in the target cities is $\leq 2$.
These results show that data scarcity does affect the performance of short-term mobility-based crime prediction.
The other lines in the plots suggest that the F1 scores of fine-tuned models tend to be higher than the \textit{baseline} model in most cases, especially when the number of months of collected data in the target city is $\leq 2$. 
When the number of months is $\geq 3$, as the number of months increases, F1 scores of all baseline models and fine-tuned models are mostly stable and closer to each other. For example, the F1 scores of all models for property crime prediction in the target city Minneapolis fall into the range of 0.58 and 0.59.

To further investigate the effects that transfer learning has on the performance of short-term crime prediction with mobility data, we compute the relative change in F1 score between the fine-tuned model $Model_{s,t}$ and the baseline model $Model_{t}$ as described previously.
Table \ref{s3-tab:transfer-ppt-f1} and \ref{s3-tab:transfer-vln-f1} show the results of relative change for all target cities using property or violent crime data.

\begin{table}[]
\centering \small
\begin{tabular}{@{}ccrrrrrrr@{}}
\toprule
Target               & Source & \multicolumn{7}{c}{Number of months of collected data in the target city} \\
city                 & city   & 1      & 2       & 3       & 4       & 5       & 6       & 7       \\
\midrule
\multirow{4}{*}{Bal} & Voting & \textbf{3.17\%} & \textbf{2.64\%}  & \textbf{1.51\%}  & \textbf{1.86\%}  & \textbf{1.87\%}  & \textbf{1.34\%}  & 1.11\%  \\
                     & Min    & 2.95\% & 2.35\%  & 1.37\%  & 1.17\%  & 0.83\%  & 1.29\%  & \textbf{1.15\%}  \\
                     & Aus    & 1.83\% & 1.65\%  & 0.74\%  & 0.53\%  & 1.46\%  & 0.43\%  & 0.58\%  \\
                     & Chi    & 2.90\% & 1.78\%  & 1.26\%  & 0.99\%  & 1.18\%  & 0.31\%  & 0.37\%  \\
\midrule
\multirow{4}{*}{Min} & Voting & \textbf{1.59\%} & 0.52\%  & \textbf{0.14\%}  & \textbf{0.57\%}  & 0.04\%  & \textbf{0.30\%}  & \textbf{-0.34\%} \\
                     & Bal    & 1.09\% & -0.45\% & -0.99\% & -0.33\% & -0.96\% & -1.06\% & -1.10\% \\
                     & Aus    & 1.42\% & \textbf{0.66\%}  & -0.37\% & -0.03\% & -0.29\% & 0.02\%  & -0.42\% \\
                     & Chi    & 1.30\% & 0.43\%  & -0.36\% & 0.09\%  & \textbf{0.13\%}  & -0.09\% & -0.71\% \\
\midrule
\multirow{4}{*}{Aus} & Voting & \textbf{1.50\%} & \textbf{1.58\%}  & \textbf{1.12\%}  & \textbf{0.51\%}  & 0.22\%  & \textbf{-0.04\%} & \textbf{0.35\%}  \\
                     & Bal    & 0.92\% & 0.69\%  & 0.76\%  & 0.08\%  & \textbf{0.26\%}  & -0.39\% & -0.16\% \\
                     & Min    & 0.40\% & 0.76\%  & 0.17\%  & -0.42\% & -0.43\% & -0.45\% & -0.35\% \\
                     & Chi    & 1.28\% & 1.25\%  & 0.55\%  & 0.19\%  & -0.03\% & -0.40\% & 0.13\%  \\
\midrule
\multirow{4}{*}{Chi} & Voting & \textbf{0.97\%} & \textbf{0.67\%}  & \textbf{0.73\%}  & \textbf{1.27\%}  & \textbf{0.26\%}  & \textbf{0.43\%}  & \textbf{0.02\%}  \\
                     & Bal    & 0.03\% & 0.24\%  & 0.30\%  & 0.78\%  & -0.26\% & -0.21\% & -0.38\% \\
                     & Min    & 0.45\% & -0.21\% & -0.04\% & 0.72\%  & -0.36\% & -0.17\% & -0.10\% \\
                     & Aus    & 0.54\% & 0.11\%  & 0.23\%  & 0.75\%  & -0.19\% & -0.08\% & -0.32\% \\
\bottomrule
\end{tabular}
\caption{Relative change in average monthly F1 score using transfer learning over all test months (Aug-Dec) for property crime prediction.}
\label{s3-tab:transfer-ppt-f1}
\end{table}

Based on the relative change in F1 score for property crime prediction in Table \ref{s3-tab:transfer-ppt-f1}, we highlight the following main observations:

1) Transfer learning is beneficial for target cities with data scarcity, especially when the number of months of available mobility data is small. As the level of data scarcity alleviates, \textit{i.e.}, the number of fine-tuning months increases, the improvement in F1 score brought by transfer learning decreases. However, F1 scores stay mostly positive in all cities and across all levels of data scarcity for the majority voting transfer approach, where the knowledge from all source cities is aggregated. In other words, when knowledge is transferred from all source cities, the crime prediction accuracy of the target city improves in most cases.

2) Looking into transfer learning from a unique source city, we observe that as the level of data scarcity alleviates, the improvement in F1 score brought by transfer learning decreases, and, in many cases, can hurt the F1 score in the crime prediction for the target cities. For example, the relative changes in F1 score are all positive when the number of fine-tuning months is $1$ and eight out of twelve of the fine-tuned models (not including \textit{Voting}) have worse F1 scores compared with the baseline when the number of fine-tuning months is $7$.
This suggests that knowledge extracted from mobility data in a single source city in the form of network parameter initialization might conflict with the local knowledge in the target city when mobility data is not that scarce. This could be due to a data distribution difference between the source city and the target city \textit{i.e.,} different mobility patterns. However, for majority voting, where the knowledge from multiple source cities is aggregated, the downside of diverse data distributions is mitigated and the relative change in F1 scores is mostly positive in all cities and across all levels of data scarcity.  

3) The effects of the transfer learning differ across different target cities. For example, the relative changes in F1 score for Baltimore (Bal) are all positive for any source city considered in this study. Also, the scale of relative changes in F1 score for Baltimore are the largest among all the four target cities across all levels of data scarcity; while Chicago benefits the least from the transfer learning of the three source cities considered. This could be because Chicago is a much larger city than the source cities and the knowledge extracted from a single small city is not enough to provide a good initialization for the model in a big target city. As shown in Table \ref{s3-tab:mobi_stats}, the number of census tracts in Chicago is 809, while the number for Baltimore, Minneapolis and Austin is 200, 116 and 204 respectively.

4) The effects of transfer learning also vary by different source cities. For example, knowledge extracted from Baltimore (Bal) has little effect ($0.03\%$) on the property crime prediction for Chicago (Chi) when the number of fine-tuning months is $1$ while knowledge from other source cities help slightly increase the F1 score ($0.45\%$ and $0.54\%$). As an approach making use of knowledge from multiple source cities, the majority voting often has the best improvement in F1 score for all four target cities, as highlighted in bold in the table. 

\begin{table}[]
\centering \small
\begin{tabular}{@{}ccrrrrrrr@{}}
\toprule
Target & Source & \multicolumn{7}{c}{Number of months of collected data in the target city}  \\
city   & city   & 1       & 2       & 3       & 4       & 5       & 6       & 7       \\
\midrule
\multirow{4}{*}{Bal} & Voting & \textbf{4.09\%}  & 0.82\%  & \textbf{1.16\%}  & \textbf{-0.04\%} & \textbf{0.63\%}  & \textbf{0.38\%}  & \textbf{0.96\%}  \\
                     & Min    & 2.79\%  & \textbf{0.93\%}  & 0.82\%  & -0.65\% & -0.17\% & -0.37\% & 0.19\%  \\
                     & Aus    & 1.69\%  & -1.04\% & 0.06\%  & -1.29\% & -1.05\% & -0.63\% & 0.02\%  \\
                     & Chi    & 3.91\%  & 0.77\%  & 0.53\%  & -0.35\% & 0.56\%  & -0.05\% & 0.42\%  \\
\midrule
\multirow{4}{*}{Min} & Voting & \textbf{4.15\%}  & \textbf{-0.18\%} & \textbf{-0.24\%} & \textbf{0.18\%}  & \textbf{-0.54\%} & \textbf{0.03\%}  & \textbf{1.54\%}  \\
                     & Bal    & 3.92\%  & -0.65\% & -1.45\% & -1.17\% & -1.31\% & -0.82\% & -0.95\% \\
                     & Aus    & -0.22\% & -2.83\% & -2.81\% & -1.75\% & -1.30\% & -1.61\% & -0.34\% \\
                     & Chi    & 3.64\%  & -1.83\% & -0.86\% & -0.83\% & -0.75\% & -1.40\% & 0.60\%  \\
\midrule
\multirow{4}{*}{Aus} & Voting & 7.42\%  & \textbf{4.88\%}  & \textbf{1.88\%}  & \textbf{3.28\%}  & \textbf{3.19\%}  & \textbf{3.06\%}  & \textbf{2.01\%}  \\
                     & Bal    & 2.51\%  & 1.82\%  & 0.32\%  & 0.19\%  & 0.68\%  & 1.70\%  & 0.41\%  \\
                     & Min    & 3.68\%  & 2.54\%  & -0.13\% & 2.39\%  & 2.29\%  & 0.78\%  & 0.36\%  \\
                     & Chi    & \textbf{9.72\%}  & 3.51\%  & -0.28\% & 1.57\%  & 2.49\%  & 1.65\%  & 1.30\%  \\
\midrule
\multirow{4}{*}{Chi} & Voting & 0.38\%  & 0.53\%  & \textbf{0.01\%}  & -0.18\% & \textbf{-0.04\%} & -0.41\% & 0.08\%  \\
                     & Bal    & -0.93\% & -0.31\% & -1.25\% & -1.25\% & -0.84\% & -1.45\% & -0.91\% \\
                     & Min    & \textbf{0.91\%}  & \textbf{1.23\%}  & -0.23\% & \textbf{0.11\%}  & -0.08\% & \textbf{0.00\%}  & \textbf{0.44\%}  \\
                     & Aus    & -1.68\% & -1.17\% & -1.19\% & -0.98\% & -0.98\% & -0.90\% & -0.80\% \\
\bottomrule
\end{tabular}
\caption{Relative change in average monthly F1 score using transfer learning over all test months (Aug-Dec) for violent crime prediction.}
\label{s3-tab:transfer-vln-f1}
\end{table}

Table \ref{s3-tab:transfer-vln-f1} shows the results for transfer learning with models trained with mobility data and violent crime data. Most of the observations for violent crime prediction are similar to the property crime prediction, including 
1) transfer learning improves the prediction accuracy when the fine-tuning data in the target cities is limited;
2) the effects of transfer learning vary across different target and source cities and Chicago is the target city which benefits the least from the transferred knowledge from the source cities considered in this study; and
3) the majority voting tends to provide the best performance compared with transfer learning from a single source city. 

However, there are some findings for violent crime prediction that are different when compared with property crime prediction:

1) The improvement in F1 score tends to be larger for violent crime prediction. For example, the largest improvement in F1 score ($9.72\%$) is observed for Austin (Aus) with knowledge transferred from Chicago (Chi) when the number of fine-tuning months is $1$. While the largest improvement in F1 score for property crime prediction is $3.71\%$.

2) The variance in relative change of F1 score among different source cities is larger for violent crime prediction than for property crime prediction. For example, the relative changes in F1 score for property crime prediction in Minneapolis (Min) as the target city are $1.09\%$, $1.42\%$ and $1.30\%$ when the source cities are Baltimore, Austin and Chicago. But the corresponding relative changes for violent crime prediction are $3.92\%$, $-0.22\%$ and $3.64\%$. This suggests that for violent crime prediction, it is important to choose an appropriate source city or develop a good mechanism to incorporate knowledge from multiple source cities.

\section{Conclusion}
Deep learning architectures enhanced with human mobility data have been shown to improve the accuracy of short-term crime prediction models trained with historical crime data.
However, human mobility data may be scarce in some regions, resulting in a lack of data necessary to 
properly train short-term crime prediction models with mobility data. 
In this paper, we have proposed a novel transfer learning framework for deep learning, short-term crime prediction models. 
Specifically, we use a network-based transfer learning approach whereby weights from the deep learning crime prediction models trained in \textit{source} cities with plenty of mobility data are transferred to \textit{target} cities to fine-tune their local crime prediction models and improve crime prediction accuracy.

Our results have shown that the proposed majority voting transfer learning - whereby weights from all source cities are transferred - improves the F1 scores for target cities with mobility data scarcity, especially when the number of months of available mobility data is small. We also reveal that these accuracy improvements are pervasive across both violent and property crimes, with larger improvements associated with violent crime prediction.
On the other hand, we have also shown that knowledge transfer from a unique source city does not work as well, with multiple cases showing decreases in F1 accuracy for target cities after the knowledge transfer. 
Finally, our results highlight that the F1 score improvements for the majority voting transfer learning approach proposed 
are not homogeneous across cities, with smaller target cities being associated with largest improvements. 

\bibliography{ref} 

\end{document}